%% file: icml.tex
\documentclass[]{article} 
\usepackage{microtype}
\usepackage{graphicx}
\usepackage{booktabs} 
\usepackage{url}
\usepackage{multirow}

\usepackage{hyperref}

\usepackage[accepted]{icml2019}

\usepackage{subcaption}
\usepackage{latexsym}
\usepackage{amsmath,amssymb} 
\usepackage{mathtools}       
\usepackage{stmaryrd}        
\usepackage{bm}              
\providecommand{\argmax}{\operatornamewithlimits{argmax}} 
\DeclareMathOperator{\Tr}{Tr}     
\DeclareMathOperator{\Cov}{Cov}   
\DeclareMathOperator{\diag}{diag} 
\providecommand{\R}{\mathbb{R}} 
\providecommand{\E}{\mathbb{E}} 
\providecommand{\T}{\mathrm{T}} 
\DeclarePairedDelimiterX{\inner}[2]{\langle}{\rangle}{#1, #2}
\DeclarePairedDelimiter{\norm}{\lVert}{\rVert}

\usepackage{amsthm}
\newtheorem{theorem}{Theorem}[]
\newtheorem{proposition}[theorem]{Proposition}

\theoremstyle{definition}

\makeatletter
\newcommand{\figcaption}[1]{\def\@captype{figure}\caption{#1}}
\newcommand{\tblcaption}[1]{\def\@captype{table}\caption{#1}}
\makeatother

\renewcommand{\paragraph}[1]{\textbf{#1}\ }

\icmltitlerunning{Adaptive Stochastic Natural Gradient Method for One-Shot Neural Architecture Search}

\begin{document}

\twocolumn[
\icmltitle{Adaptive Stochastic Natural Gradient Method for One-Shot Neural Architecture Search}



\icmlsetsymbol{equal}{*}

\begin{icmlauthorlist}
\icmlauthor{Youhei Akimoto}{equal,tsukuba}
\icmlauthor{Shinichi Shirakawa}{equal,yokohama}
\icmlauthor{Nozomu Yoshinari}{yokohama}
\icmlauthor{Kento Uchida}{yokohama}
\icmlauthor{Shota Saito}{yokohama,kaisha}
\icmlauthor{Kouhei Nishida}{shinshu}
\end{icmlauthorlist}

\icmlaffiliation{tsukuba}{University of Tsukuba \& RIKEN AIP}
\icmlaffiliation{yokohama}{Yokohama National University}
\icmlaffiliation{kaisha}{SkillUp AI Co., Ltd.}
\icmlaffiliation{shinshu}{Shinshu University}

\icmlcorrespondingauthor{Youhei Akimoto}{akimoto@cs.tsukuba.ac.jp}
\icmlcorrespondingauthor{Shinichi Shirakawa}{shirakawa-shinichi-bg@ynu.ac.jp}

\icmlkeywords{Neural Architecture Search, Stochastic Relaxation, Stochastic Natural Gradient, Step-Size Adaptation, Theoretical Guarantee}

\vskip 0.3in
]



\printAffiliationsAndNotice{\icmlEqualContribution} 

\begin{abstract}
High sensitivity of neural architecture search (NAS) methods against their input such as step-size (i.e., learning rate) and search space prevents practitioners from applying them out-of-the-box to their own problems, albeit its purpose is to automate a part of tuning process. Aiming at a fast, robust, and widely-applicable NAS, we develop a generic optimization framework for NAS. We turn a coupled optimization of connection weights and neural architecture into a differentiable optimization by means of stochastic relaxation. It accepts arbitrary search space (widely-applicable) and enables to employ a gradient-based simultaneous optimization of weights and architecture (fast). We propose a stochastic natural gradient method with an adaptive step-size mechanism built upon our theoretical investigation (robust). Despite its simplicity and no problem-dependent parameter tuning, our method exhibited near state-of-the-art performances with low computational budgets both on image classification and inpainting tasks.
\end{abstract}

\newcommand{\xx}{\bm{x}}
\newcommand{\X}{\mathcal{X}}
\newcommand{\cc}{\bm{c}}
\newcommand{\C}{\mathcal{C}}
\newcommand{\Fi}{\mathbf{F}}

\section{Introduction}

Neural architecture search (NAS) is a promising way to automatically find a reasonable neural network architecture and one of the most popular research topics in deep learning. The success of deep learning impresses people from outside machine learning communities and attracts practitioners to apply deep learning to their own tasks. However, they face different difficulties when applying deep learning. One difficulty is to determine the neural architecture for their previously-unseen problem. NAS is a possible solution to this difficulty. 

Work published before 2017 often frames NAS as a hyper-parameter optimization, where an architecture's performance is measured by the validation error obtained after the training of the weights under a fixed architecture \citep{Real2017,Suganuma2017,Zoph2017}. More recent studies \citep{Brock2018,Shirakawa2018,Pham2018,Liu2019,Xie2019,Cai2019}, on the other hand, optimize the weights and the architecture simultaneously within a single training by treating all possible architectures as subgraphs of a supergraph. These approaches are called \emph{one-shot architecture search} or \emph{one-shot NAS}. They break through the bottleneck of the hyper-parameter optimization approaches, namely, high computational cost for each architecture evaluation, and enable to perform NAS on a standard personal computer, leading to gathering more potential applications.

Research directions of NAS fall into three categories \citep{Elsken2019}: performance estimation (how to estimate the performance of architectures), search space definition (how to define the possible architectures), and search strategy (how to optimize the architecture). In the last direction, promising approaches transform a coupled optimization of weights and architectures into optimization of a differentiable objective by means of \emph{continuous relaxation} \cite{Liu2019,Xie2019} or \emph{stochastic relaxation} \cite{Shirakawa2018,Pham2018}. A gradient descent or a natural gradient descent strategy with an existing adaptive step-size mechanism or a constant step-size is then employed to optimize weights and architecture simultaneously. However, optimization performance is sensitive against its inputs such as step-size (i.e., learning rate) and search space, limiting its application to unseen tasks. 

To achieve a robust NAS, we develop a generic optimization framework for one-shot NAS. Our strategy is based on stochastic relaxation. We generalize the work by \citet{Shirakawa2018} to enable arbitrary types of architecture variables including categorical variables, ordinal (such as real or integer) variables, and their mixture. We develop a unified optimization framework for our stochastic relaxation based on the so-called stochastic natural gradient \cite{Amari1998}. Our theoretical investigation derives a condition on the step-size for our objective value to improve monotonically every iteration. We propose a step-size adaptation mechanism to approximately satisfy the condition. It significantly relaxes the performance sensitivity on the inputs and makes the overall framework rather flexible. 

Our contributions are summarized as follows: (i) Our framework can treat virtually arbitrary types of architecture variables as long as one can define a parametric family of probability distributions on it; (ii) We propose a step-size adaptation mechanism for the stochastic natural gradient ascent, improving the optimization speed as well as its robustness against the hyper-parameter tuning. The default values are prepared for all introduced hyper-parameters, and they need not be touched even when the architecture search space changes; (iii) The proposed approach can enjoy parallel computer architecture, while it is comparable or even faster than existing approaches even on serial implementation; and (iv) Our strategy is rather simple, allowing us theoretical investigation, based on which we develop the step-size adaptation mechanism.

\section{Our Approach}

In this paper we address the following optimization problem
\begin{equation}
  \max_{\xx \in \X,\ \cc \in \C}f(\xx, \cc) \enspace,
  \label{eq:problem}
\end{equation}
where $f : \X \times \C \to \R$ is the objective function that is differentiable with respect to~(w.r.t.) $\xx \in \X$ and is black-box w.r.t.~$\cc \in \C$. The domain $\X$ of $\xx$ is a subset of $n_x$ dimensional real space $\R^{n_x}$, whereas $\C$ can be either categorical, continuous, or their product space. Our objective is to simultaneously optimize $\xx$ and $\cc$ by possibly utilizing the gradient $\nabla_{\xx} f$. In the context of NAS, $\xx$, $\cc$ and $f$ represent connection weights, architecture parameters, a criterion to be maximized such as negative loss.

\subsection{Stochastic Relaxation}

We turn the original optimization problem into an optimization of differentiable objective $J$ by means of \emph{stochastic relaxation}. For this purpose, we introduce a family of probability distributions $\mathcal{P} = \{P_\theta : \theta \in \Theta \subseteq \R^{n_\theta}\}$ defined on $\C$. We suppose that for any $\cc \in \C$ the family of probability distribution contains a sequence of the distributions that approaches the Dirac-Delta distribution $\delta_{\cc}$ concentrated at $\cc$. Moreover, we suppose that any $P_\theta \in \mathcal{P}$ admits the density function $p_\theta$ w.r.t.\ the reference measure $d\cc$ on $\C$, and the log-density is differentiable w.r.t.\ $\theta \in \Theta$. The stochastic relaxation of $f$ given $\mathcal{P}$ is defined as follows
\begin{equation}
  J(\xx, \theta) := \int_{\cc \in \C} f(\xx, \cc) p_\theta(\cc) d\cc = \E_{p_\theta}[f(\xx, \cc)]\enspace.
  \label{eq:SR}
\end{equation}
Maximization of $J$ coincides with maximization of $f$, as $\sup_{\theta \in \Theta} J(\xx, \theta) = \sup_{\cc \in \C} f(\xx, \cc) = f(\xx, \cc^*)$, where the supremum of $J(\xx, \theta)$ is attained by the limit of the sequence $\{\theta\}$ where $P_{\theta}$ converges to $\delta_{\cc^*}$.

The stochastic relaxation $J$ inherits nice properties of $f$. For example, if $f(\xx, \cc)$ is convex and/or Lipschitz continuous w.r.t.~$\xx$, then so is $J(\xx, \theta)$ w.r.t.~$\xx$, respectively. Moreover, the stochastic relaxation $J$ is differentiable w.r.t.~both $\xx$ and $\theta$ under mild conditions as follows
\begin{align}
  \nabla_{\xx} J(\xx, \theta) &= \E_{p_\theta}[\nabla_{\xx} f(\xx, \cc)] \label{eq:jx}\\
  \nabla_{\theta} J(\xx, \theta) &= \E_{p_\theta}[f(\xx, \cc) \nabla_{\theta}\ln(p_\theta(\cc)) ] \label{eq:jt}  \enspace.
\end{align}

\paragraph{Stochastic Relaxation with Exponential Family:}
An exponential family consists of probability distributions whose density is expressed as $h(\cc) \cdot \exp( \eta(\theta)^\T T(\cc) - \varphi(\theta) )$, where $T: \C \to \mathbb{R}^{n_\theta}$ is the sufficient statistics, $\eta: \Theta \to \mathbb{R}^{n_\theta}$ is the natural parameter of this family, and $\varphi(\theta)$ is the normalization factor. For the sake of simplicity, we limit our focus on the case $h(\cc) = 1$. If we choose the parameter $\theta$ so that $\theta = \E_{p_\theta}[T(\cc)]$, it is called the \emph{expectation parameters} of this family. Under the expectation parameters, the natural gradient of the log-likelihood reduces to $\tilde \nabla \ln(p_\theta(\cc)) = T(\cc) - \theta$. The inverse of Fisher information matrix is $\Fi^{-1} (\theta) = \mathbb{E}[(T(\cc) - \theta)(T(\cc) - \theta)^\mathrm{T}]$, and is typically expressed as an analytical function of $\theta$.

\subsection{Alternating Gradient Ascent}

Let $\xx^t$ and $\theta^t$ represent the parameter values at iteration $t$. We maximize \eqref{eq:SR} by alternating optimization, namely,
\begin{align}
  \xx^{t+1} &= \textstyle \argmax_{x \in \X}\ J(\xx, \theta^t)
  \label{eq:xt}\\
  \theta^{t+1} &= \textstyle \argmax_{\theta \in \Theta}\ J(\xx^{t+1}, \theta) \enspace. \label{eq:tt}
\end{align}
Alternating steepest ascent is a way to avoid repeatedly solving computationally heavy optimization \eqref{eq:xt} and \eqref{eq:tt}. 

\paragraph{Stochastic Gradient Ascent on $\X$:}
Update step \eqref{eq:xt} is replaced with the gradient ascent w.r.t.~a metric $\mathbf{A}$ on $\X$ with possibly time-dependent step-size $\epsilon_{\xx}$,
\begin{align}
  \xx^{t+1} = \xx^t + \epsilon_{\xx} \mathbf{A} \nabla_{\xx} J(\xx^t, \theta^t) \label{eq:gg} \enspace.
\end{align}
Here $\mathbf{A}$ may change over $t$, leading to (quasi-) second order update. For fixed $\theta^{t}$, it has been widely investigated in literature, and convergence of $\xx$ to a stationary point $\norm{\nabla_{\xx} J(\xx, \theta^{t})} = 0$ is guaranteed under different conditions.

Monotone improvement of $J$ is easily derived under different conditions. An example result is as follows.
\begin{proposition}\label{prop:xx}
  Assume that $J(\xx, \theta^{t})$ is $\norm{\cdot}_{\mathbf{A}}$-Lipschitz smooth w.r.t.~$\xx$: $\norm{\nabla_{\xx} J(\xx', \theta^t) - \nabla_{\xx} J(\xx, \theta^t)}_{\mathbf{A}} \leq L \norm{\xx' - \xx}_{\mathbf{A}}$. (This is satisfied if $f(\xx, \cc)$ is so for all $\cc$.) Then, for $\epsilon_{\xx} < 2/L$, we have the monotone improvement
   \begin{multline}
     J(\xx^{t+1}, \theta^t) - J(\xx^{t}, \theta^t) \\
     \geq (\epsilon_{\xx} - (L/2)\epsilon_{\xx}^2) \norm{\nabla_{\xx} J(\xx^{t}, \theta^t)}_{\mathbf{A}}^2 > 0 \enspace.
   \end{multline}
\end{proposition}

In our situation, the gradient $\nabla_{\xx}J(\xx^{t}, \theta^{t})$ is not tractable. Instead, we estimate it by Monte-Carlo (MC) using $\nabla_{\xx}J(\xx^{t}, \cc_i)$ with independent and identically distributed (i.i.d.) samples $\cc_i \sim P_{\theta^t}$ ($i = 1,\dots, \lambda_{\xx}$), namely, 
\begin{equation}
   G_{\xx}(\xx^t, \theta^t) = \frac{1}{\lambda_{\xx}} \sum_{i=1}^{\lambda_{\xx}} \nabla_{\xx} f(\xx^{t}, \cc_i) \enspace.\label{eq:gx}
\end{equation}
The strong law of large numbers shows $\lim_{\lambda_{\xx} \to \infty}G_{\xx}(\xx^t, \theta^t) = \nabla_{\xx} J(\xx^t, \theta^t)$ almost surely under mild conditions. The number $\lambda_{\xx}$ of MC samples determines the trade-off between the accuracy and the computational cost.

We replace $\nabla_{\xx} J(\xx^t, \theta^t)$ with $G_{\xx}(\xx^t, \theta^t)$, leading to a stochastic gradient ascent, for which adaptation mechanisms for the step-size $\epsilon_{\xx}$ are developed.

\paragraph{Stochastic Natural Gradient Ascent on $\Theta$:}
Update step \eqref{eq:tt} is replaced with the natural gradient ascent with gradient normalization and step-size $\epsilon_{\theta}$,
\begin{align}
  \theta^{t+1} &= \theta^t + \epsilon_{\theta}  \tilde\nabla_{\theta} J(\xx^t, \theta^t) \label{eq:ng}\\
  \epsilon_{\theta} &= \delta_{\theta} / \norm{ \tilde\nabla_{\theta} J(\xx^t, \theta^t) }_{\Fi(\theta^t)}\label{eq:nge} \enspace,
\end{align}
where $\tilde \nabla_{\theta} = \Fi(\theta^t)^{-1} \nabla_{\theta}$. 
It can be approximately understood as the trust region method under the Kullback-Leibler (KL-) divergence with trust region radius $\delta_\theta$.

As the natural gradient is not analytically obtained, we use MC to obtain its approximation
\begin{equation}
  G_\theta(\xx^{t+1}, \theta^t) = \frac{1}{\lambda_{\theta}} \sum_{i=1}^{\lambda_{\theta}} f(\xx^{t+1}, \cc_i) (T(\cc_i) - \theta^t) \enspace,\label{eq:gt}
\end{equation}
where $\cc_i$ are i.i.d.~from $P_{\theta^t}$.
The parameter update follows
\begin{align}
  \theta^{t+1} &= \theta^t + \epsilon_{\theta} G_\theta(\xx^{t+1}, \theta^t) \label{eq:sng}\\
  \epsilon_\theta &= \delta_{\theta} / \norm{G_\theta(\xx^{t+1}, \theta^t)}_{\Fi(\theta^t)} \enspace.
\end{align}

\subsection{Adaptive Stochastic Natural Gradient}

In general, the step-size of a stochastic gradient algorithm plays one of the most important roles in performance and optimization time. Different adaptive step-size mechanisms have been proposed such as Adam \citep{Kingma2015}.
However, our preliminary empirical study shows a specific adaptation mechanism for $\epsilon_\theta$ is required to have robust performance.
In the following, we first investigate the theoretical properties of the stochastic natural gradient ascent introduced above, then we introduce an adaptation mechanism for the trust-region radius $\delta_{\theta}$.

\paragraph{Theoretical Background:}
For problems without $\xx$, i.e., fully black-box optimization of $f(\cc)$, the natural gradient ascent \eqref{eq:ng} of the stochastic relaxation \eqref{eq:SR} of function $f(\cc)$ is known as the \emph{information geometric optimization} (IGO) \cite{Ollivier2017} algorithm. For the case of exponential family with expectation parameters, \citet{AkimotoFOGA2013} have shown that \eqref{eq:ng} leads to a monotone increase of $J(\theta)$, summarized as follows.\footnote{The statement is simplified so as not to introduce additional notation. Note that if $f(\cc)$ is lower bounded, considering $f(\cc) - \min_{\cc \in \C} f(\cc)$ in \eqref{eq:SR} instead of $f$ is sufficient to meet the condition of Proposition~\ref{thm:qi}. This modification only adds an offset to the $J$ value without affecting the gradient.}
\begin{proposition}[Theorem~12 of \cite{AkimotoFOGA2013}]\label{thm:qi}
  Assume that $\min_{\cc \in \C} f(\cc) > 0$.  Then, \eqref{eq:ng} satisfies
  \begin{multline}
    \ln J(\theta^{t}+\epsilon_{\theta}  \tilde\nabla_{\theta} J(\theta^t)) - \ln J(\theta^{t}) \\
    \geq  ((\epsilon_{\theta} J(\theta^t))^{-1} - 1) D_\theta (\theta^{t}+\epsilon_{\theta}  \tilde\nabla_{\theta} J(\theta^t), \theta^t)\enspace,
  \end{multline}
\end{proposition}
where $D_{\theta}$ is KL-divergence on $\Theta$.

Proposition~\ref{thm:qi} gives us a very useful insight into the step-size $\epsilon_\theta$. It says that $\epsilon_\theta < 1/J(\theta^t)$ leads to improvement in $J$ value as long as the parameter follows the \emph{exact} natural gradient. Together with Proposition~\ref{prop:xx}, it implies the monotone improvement of alternating update of $\xx$ and $\theta$ when the exact gradients are given. However, in our situation, the natural gradient in \eqref{eq:ng} is not tractable and one needs to approximate it with MC. Then, the monotone improvement is not guaranteed. A promising feature of our framework is that the MC approximate $G_\theta(\xx^{t+1}, \theta^t)$ of the natural gradient $\tilde\nabla_\theta J(\theta^t)$ can be made arbitrarily accurate by taking the number of MC samples $\lambda_{\theta}$ to $\infty$. 

The following proposition shows that the loss in $J$ is bounded if the divergence is bounded. Its proof can be found in supplementary material.
\begin{proposition}\label{thm:jloss}
  Assume that $\min_{\cc \in \C} f(\cc) > 0$ and let $f^* = \max_{\cc \in \C} f(\cc)$. Then, $\ln J(\theta') - \ln J(\theta) \geq - \frac{f^*}{J(\theta)} D_\theta(\theta', \theta)$ for any $\theta$ and $\theta'$.
\end{proposition}

As a straight-forward consequence of the above two propositions, we obtain a sufficient conditions for the stochastic natural gradient ascent to improve $J$ monotonically. This is the baseline of our proposal.
\begin{theorem}\label{thm:main}
  Assume that $\min_{\cc \in \C} f(\cc) > 0$ and let $f^* = \max_{\cc \in \C} f(\cc)$.
  For any $\epsilon > 0$, 
  if $D_\theta(\theta, \theta^t + \epsilon \tilde\nabla_\theta J(\theta^t) ) \leq \zeta D_\theta(\theta^{t}+ \epsilon \tilde\nabla_{\theta} J(\theta^t), \theta^t)$ holds for some $\zeta > 0$, we have
  \begin{multline}
    \ln J(\theta) - \ln J(\theta^{t}) \\
    \geq \frac{1 - \zeta \epsilon f^* - \epsilon J(\theta^t)}{\epsilon J(\theta^t)} D_\theta(\theta^{t}+ \epsilon \tilde\nabla_{\theta} J(\theta^t), \theta^t) \enspace.
  \end{multline}
  In particular, if $\epsilon < (\zeta f^* + J(\theta^t))^{-1}$ holds, $J(\theta) > J(\theta^{t})$. 
\end{theorem}
If we replace $\theta$ with $\theta^{t+1}$ defined in \eqref{eq:sng}, we obtain a sufficient condition for the stochastic natural gradient update \eqref{eq:sng} to lead to monotone improvement, namely,
\begin{multline}
  D_\theta(\theta^{t+1}, \theta^t + \epsilon_\theta \tilde\nabla_\theta J(\xx^{t+1}, \theta^t) ) \\
  \leq \zeta D_\theta(\theta^{t}+ \epsilon_{\theta}  \tilde\nabla_{\theta} J(\xx^{t+1}, \theta^t), \theta^t)\enspace.
  \label{eq:maincond}
\end{multline}
This can be satisfied for any $\zeta > 0$ by taking a sufficiently large $\lambda_{\theta}$ as $G_\theta(\xx^{t+1}, \theta^t)$ is a consistent estimator of $\tilde\nabla_{\theta} J(\xx^{t+1}, \theta^t)$ and the left hand side~(LHS) is $O(\lambda_{\theta}^{-1})$.

However, if $\epsilon_\theta$ (or $\delta_{\theta}$) is sufficiently small, monotone improvement at each iteration is too strict and one might only need to guarantee the improvement over $\tau > 0$ iterations, where $\tau \propto 1/\delta_{\theta}$. To derive an insightful formula, we put aside the mathematical rigor in the following. Let $\tilde\nabla_{\lambda_\theta}^t = G_\theta(\xx^{t+1}, \theta^t)$ for short.
We continue to consider a problem without $\xx$ (or $\xx$ is fixed).
A common argument borrowed from stochastic approximation (e.g., \citet{Borkar2008book}) states that
if $\epsilon_\theta$ is so small that the parameter vector stays near $\theta^t$ and $ \tilde\nabla_{\lambda_\theta}^{t+i}$ are considered i.i.d.~for $i = 0, \dots, \tau-1$, the parameter vector after $\tau$ steps will be approximated as
\begin{equation*}
  \theta^{t + \tau} - \theta^{t} \approx  \epsilon_\theta \tau \E[\tilde\nabla_{\lambda_\theta}^t]
  +  \epsilon_\theta \tau \sum_{i=0}^{\tau-1}\frac{1}{\tau} (\tilde\nabla_{\lambda_\theta}^{t + i} - \E[\tilde\nabla_{\lambda_\theta}^t])\enspace.
\end{equation*}
If we replace $\theta^{t+1}$ with $\theta^{t+\tau}$ and $\epsilon$ with $\tau \epsilon_\theta$ in \eqref{eq:maincond} and apply the approximation of the KL-divergence by the Fisher information matrix, we obtain
\begin{equation*}
 \underbrace{\norm*{\sum_{i=0}^{\tau-1}\frac{\tilde\nabla_{\lambda_\theta}^{t + i} - \E[\tilde\nabla_{\lambda_\theta}^t]}{\sqrt{\tau}}}_{\Fi(\theta^t)}^2}_{\to \Tr(\Cov(\tilde\nabla_{\lambda_\theta}^{t}) \Fi(\theta^t)) \ \text{as}\ \tau \to \infty}
\leq \zeta \tau \norm{ \E[\tilde\nabla_{\lambda_\theta}^t] }_{\Fi(\theta^t)}^2\enspace,
\end{equation*}
The LHS tends to the variance of $\tilde\nabla_{\lambda_\theta}^t$ measured w.r.t.~the Fisher metric and is upper bounded by $(f^*)^2 n_\theta /\lambda_{\theta}$. That is, $\lambda_{\theta}$ and/or $\delta_{\theta}$ should be adapted so that
\begin{equation}
 \frac{ \norm{ \E[ \tilde\nabla_{\lambda_{\theta}}^t] }_{\Fi(\theta^t)}^2 }{ \Tr(\Cov(\tilde\nabla_{\lambda_\theta}^{t}) \Fi(\theta^t)) } \geq \frac{1}{ \zeta \tau }  \in \Omega(\delta_\theta) \enspace.
  \label{eq:snr}
\end{equation}
In words, the signal-to-noise ratio (LHS of \eqref{eq:snr}) must be greater than a constant proportional to $\delta_\theta$. 

\paragraph{Adaptive Stochastic Natural Gradient:}
We develop an algorithm that approximately satisfies the above-mentioned condition by adapting the trust region $\delta_{\theta}$. The above condition can be satisfied by increasing $\lambda_{\theta}$ while $\delta_{\theta}$ is fixed, and the same idea as described below can be used to adapt the number $\lambda_\theta$ of MC samples. The reason we adapt $\delta_{\theta}$ rather than $\lambda_{\theta}$ is to update connection weights $\xx$ more frequently ($\xx$ is updated after every $\lambda_\theta$ forward network processes). If multiple GPUs are available, one can set $\lambda_{\theta} = \lambda_{\xx} = \#\text{GPUs}$ and enjoy parallel computation, allowing to keep $\epsilon_{\xx}$ and $\delta_{\theta}$ (hence $\epsilon_{\theta}$ as well) higher as the stochastic gradient becomes more reliable.

We introduce the accumulation of the stochastic natural gradient as follows
\begin{align}
  \bm{s}^{(t+1)} &= (1 - \beta) \bm{s}^{(t)} + \sqrt{\beta(2 - \beta)} \Fi(\theta^{t})^{\frac12} \tilde\nabla_{\lambda_{\theta}}^{t} \enspace,\label{eq:s} \\
  \gamma^{(t+1)} &= (1 - \beta)^2 \gamma^{(t)} + \beta(2 - \beta) \norm{\tilde\nabla_{\lambda_{\theta}}^{t}}_{\Fi(\theta^{t}) }^2 \enspace,\label{eq:g}
\end{align}
where $\bm{s}^{(0)} = \bm{0}$ and $\gamma^{(0)} = 0$.
To understand the effect of $\bm{s}$ and $\gamma$, we consider the situation that $\epsilon_{\xx}$ and $\delta_\theta$ are small enough that $\xx^{t}$ and $\theta^{t}$ stay at $(\xx, \theta)$. 
Then, $\bm{s}^t$ approaches $\sqrt{(2-\beta)/\beta} \E[\Fi(\theta^t)^\frac12\tilde \nabla_{\lambda_\theta}] + \xi$, where $\xi$ is a random vector with $\E[\xi] = \bm{0}$ and $\Cov(\xi) = \Fi(\theta^t)^\frac12 \Cov(\tilde \nabla_{\lambda_\theta})\Fi(\theta^t)^\frac12$, and $\gamma^t$ approximates $\E[\norm{\tilde \nabla_{\lambda_\theta}}_{\Fi(\theta^t)}^2] = \norm{\E[\tilde \nabla_{\lambda_\theta}]}_{\Fi(\theta^t)}^2 + \Tr(\Cov(\tilde\nabla_{\lambda_\theta}^{t}) \Fi(\theta^t))$. If we set $\beta \propto \delta_\theta$ and adapt $\lambda_{\theta}$ or $\delta_{\theta}$ to keep $\norm{\bm{s}^{(t+1)}}^2 / \gamma^{(t+1)} \geq \alpha$ for some $\alpha > 1$, it approximately achieves
\begin{multline*}
  \frac{\norm{\E[\tilde \nabla_{\lambda_\theta}]}_{\Fi(\theta^t)}^2}{\Tr(\Cov(\tilde\nabla_{\lambda_\theta}^{t}) \Fi(\theta^t))}
  \geq \frac{\norm{\E[\tilde \nabla_{\lambda_\theta}]}_{\Fi(\theta^t)}^2}{\E[\norm{\tilde \nabla_{\lambda_\theta}}_{\Fi(\theta^t)}^2]} \\
  \approx \frac{\beta}{2 - 2\beta}\left(\frac{\norm{\bm{s}^{(t+1)}}^2}{ \gamma^{(t+1)} } - 1\right)
  \geq \frac{\beta(\alpha - 1)}{2 - 2\beta} \in \Theta(\delta_\theta) 
  \enspace.
\end{multline*}
It results in satisfying \eqref{eq:snr}. 

The adaptation of $\delta_{\theta}$ is then done as follows:  
\begin{equation}
  \delta_{\theta} \leftarrow \delta_{\theta} \exp\left( \beta \left( \norm{\bm{s}^{(t+1)}}^2 / \alpha - \gamma^{(t+1)} \right) \right) \enspace.
\end{equation}
This tries to keep $\norm{\bm{s}^{(t+1)}}^2 / \gamma^{(t+1)} \approx \alpha$ by adapting $\delta_{\theta}$.

\newcommand*{\proposed}[1][true]{\ifthenelse{\boolean{#1}}{ASNG-NAS\ }{ASNG-NAS}} 

\subsection{Adaptive Stochastic Natural Gradient-based NAS}
\label{sec:asngnas}
 \begin{algorithm}[t]\caption{\proposed[false]}\label{algo:pdas}
 \begin{algorithmic}[1]
   \REQUIRE{$\xx^0$, $\theta^0$} \COMMENT{initial search points} 
   \REQUIRE{$\alpha = 1.5$, $\delta_\theta^{0}  = 1$, $\lambda_{\xx} = \lambda_{\theta} = 2$}
   \STATE $\Delta = 1$, $\gamma = 0$, $\bm{s} = \bm{0}$, $t = 0$
   \REPEAT
   \STATE $\delta_{\theta} = \delta_{\theta}^0 / \Delta$, $\beta = \delta_{\theta} / n_{\theta}^{1/2}$
   \STATE compute $G_{\xx}(\xx^t, \theta^t)$ by \eqref{eq:gx} and update $\xx^{t+1}$ using $G_{\xx}(\xx^t, \theta^t)$
   \STATE compute $G_{\theta}(\xx^{t+1}, \theta^t)$ by \eqref{eq:gt}, update $\theta^{t+1}$ with \eqref{eq:sng}, then force $\theta^{t+1} \in \Theta$ by projection
   \STATE $\bm{s} \gets (1 - \beta) \bm{s} + \sqrt{\beta(2 - \beta)} \frac{\Fi(\theta^{t})^{\frac12} G_{\theta}(\xx^{t+1}, \theta^t)}{ \norm{G_{\theta}(\xx^{t+1}, \theta^t)}_{\Fi(\theta^t)} }$
   \STATE $\gamma \gets (1 - \beta)^2 \gamma + \beta(2 - \beta)$
   \STATE $\Delta \gets  \min(\Delta_{\max}, \Delta \exp( \beta ( \gamma - \norm{\bm{s}}^2 / \alpha )))$
   \UNTIL{termination conditions are met}
 \end{algorithmic}
 \end{algorithm}

The proposed optimization method for problem \eqref{eq:problem}, called Adaptive Stochastic Natural Gradient-based NAS (\proposed[false]), is summarized in Algorithm~\ref{algo:pdas}.
Here, we summarize some implementation remarks. One is that instead of accumulating $\Fi(\theta^{t})^{\frac12} \tilde\nabla_{\lambda_{\theta}}^{t}$ and $\norm{\tilde\nabla_{\lambda_{\theta}}^{t}}_{\Fi(\theta^{t})}^2$ separately in \eqref{eq:s} and \eqref{eq:g}, we accumulate $\Fi(\theta^{t})^{\frac12} \tilde\nabla_{\lambda_{\theta}}^{t} / \norm{\tilde\nabla_{\lambda_{\theta}}^{t}}_{\Fi(\theta^{t})}$ in $\bm{s}$ and $\gamma \approx 1$. In our preliminary experiments, we found it more stable. The other point is that the average function value is subtracted from the function value in the stochastic natural gradient computation \eqref{eq:gt} when $\lambda_{\theta} = 2$. This is a well-known technique to reduce the estimation variance of gradient while the expectation is unchanged (e.g., \citet{Evans2000book}). Since we
normalize the stochastic natural gradient when the parameter is updated, it is equivalent to transform $f_1 = f(\xx^{t+1}, \cc_{1})$ and $f_2 = f(\xx^{t+1}, \cc_{2})$ to ($1$, $-1$) if $f_1 > f_2$, ($-1$, $1$) if $f_1 < f_2$, and ($0$, $0$) if $f_1 = f_2$ (in this case, we skip the update and start the next iteration). When $\lambda_{\theta} > 2$, we similarly transform $f_i = f(\xx^{t+1}, \cc_{i})$ in \eqref{eq:gt} to $1$ if $f_i$ is in top $\lceil\lambda_{\theta}/4\rceil$, $-1$ if it is in bottom $\lceil\lambda_{\theta}/4\rceil$, and $0$ otherwise. By doing so, we obtain invariance to a strictly increasing transformation of $f$, and we observed significant speedup in many cases in our preliminary study.

To instantiate \proposed[false], we prepare an exponential family defined on $\mathcal{C}$. If $\mathcal{C}$ is a set of categorical variables ($\mathcal{C} = \llbracket 1,m_1\rrbracket\times \cdots \times\llbracket1, m_{n_{\cc}}\rrbracket$), one can simply use categorical distribution parameterized by the probability $[\theta]_{i,j} = [\theta_i]_j$ of $i$-th categorical variable to be $j$ ($1 - \sum_{j=1}^{m_i-1} [\theta]_{i,j}$ is the probability of $[\cc]_{i} = m_i$). Then, $T(\cc) = (T_1([\cc]_1), \dots, T_{n_{\cc}}([\cc]_{n_{\cc}}))$, where $T_i: \llbracket 1, m_{i}\rrbracket \to [0, 1]^{m_i-1}$ is the one-hot representation without the last element, and $\Fi(\theta) = \diag(\Fi_1(\theta_1), \dots, \Fi_{n_{\cc}}(\theta_{n_{\cc}}))$, where $\Fi_i(\theta_{i}) = \diag(\theta_{i})^{-1} + (1 - \sum_{j=1}^{m_i-1} [\theta_{i}]_{j})^{-1}\bm{1}\bm{1}^\T$. 
If $\mathcal{C}$ is a set of ordinal variables, e.g., $\mathcal{C} \subseteq \mathbb{R}^{n_{\cc}}$, our choice will be $P_\theta = \mathcal{N}(\mu_1, \sigma_1^2) \times \cdots \times\mathcal{N}(\mu_{n_{\cc}}, \sigma_{n_{\cc}}^2)$ and $\theta = (\mu_1, \mu_1^2 + \sigma_1^2, \dots, \mu_{n_{\cc}}, \mu_{n_{\cc}}^2 + \sigma_{n_{\cc}}^2)$. Then, we have $T(\cc) = (T_1([\cc]_{1}),\dots, T_{n_{\cc}}([\cc]_{n_{\cc}}))$ with $T_i([\cc]_i) = ([\cc]_{i}, [\cc]_{i}^2)$, and $\Fi(\theta)$ is a block-diagonal matrix with block size $2$ whose $i$-th block is $[\sigma_i^2, 2\mu_i\sigma_i^2; 2\mu_i\sigma_i^2, 4\mu_i^2\sigma_i^2 + 2\sigma_i^4]^{-1}$. Integer variables can be treated similarly. If $\mathcal{C}$ is a product of categorical and ordinal variable spaces, we can use their product distribution. A desired $\theta^0$ realizes the maximal entropy in $\Theta$ unless one has a prior knowledge. Moreover, $\Theta$ should be restricted to avoid degenerated distribution. E.g., for categorical distribution, we lower bounds $[\theta]_{i, j}$ by $\theta_{i}^{\min} = (n_{\cc} (m_i - 1))^{-1}$. See the supplementary material.

\newcommand{\Dataset}{\mathcal{D}}
\newcommand{\Loss}{\mathcal{L}}
\newcommand{\Obj}{\mathcal{G}}
\newcommand{\ngrad}{\tilde{\nabla}}
\newcommand{\rk}{\mathrm{rank}}
\newcommand{\wrt}{w.r.t.\ }

\section{Experiments and Results}
We investigate the robustness of ASNG on an artificial test function in \S\ref{sec:test_func}. We then apply \proposed to the architecture search for image classification and inpainting in \S\ref{sec:exp_class} and \S\ref{sec:exp_inpaint}. To compare the quality of the obtained architecture and the computational cost, we adopt the same search spaces as in previous works. The experiments were done with a single NVIDIA GTX 1080Ti GPU, and \proposed[false] is implemented using \texttt{PyTorch 0.4.1} \citep{Paszke2017}. The code is available at \url{https://github.com/shirakawas/ASNG-NAS}.

\subsection{Toy Problem}

\begin{figure}[t]
  \centering%
  \begin{subfigure}{0.49\hsize}%
    \centering%
    \includegraphics[width=\hsize]{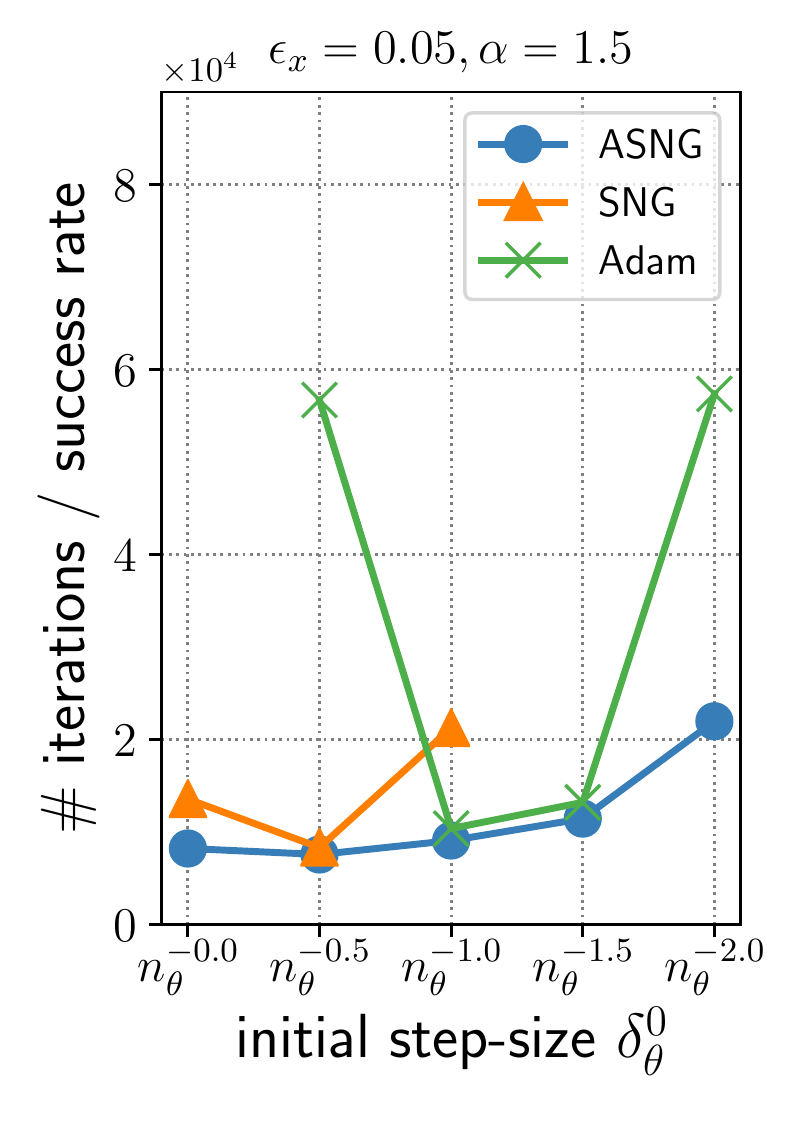}%
  \end{subfigure}%
  \begin{subfigure}{0.49\hsize}%
    \centering%
    \includegraphics[width=\hsize]{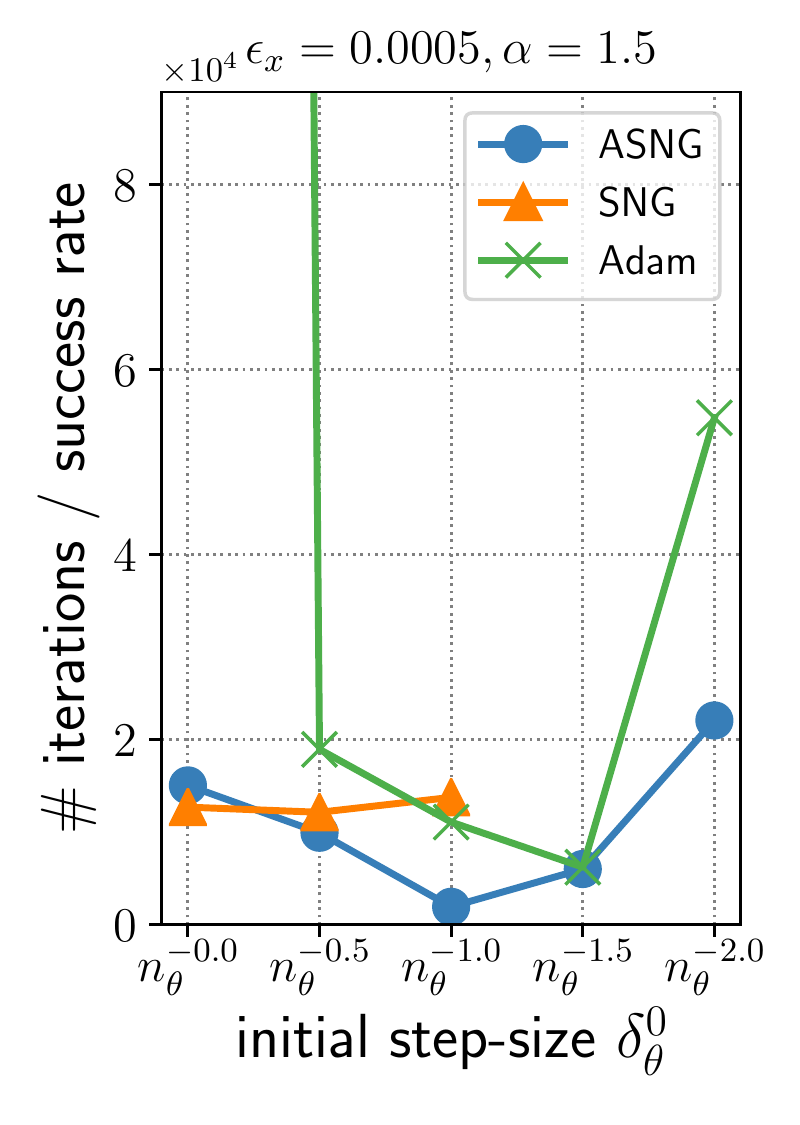}%
  \end{subfigure}%
  \caption{Results on the selective squared error function for $\epsilon_{\xx}=0.05$ and $0.0005$. Median values over $100$ runs are reported. Missing values indicate the setting never succeeded.}%
  \label{fig:benchmark}%
\end{figure}

\label{sec:test_func}
We consider the following \emph{selective squared error function} composed of continuous variable $\xx \in \R^{D \times K}$ and categorical variables $\cc$. For each $i$ ($1 \leq i \leq D$), we denote the one-hot vector of $i$-th categorical variable in $\cc$ by $h_i(\cc) \in \{0, 1\}^K$. The objective function to be minimized is
\begin{equation*}
f(\xx, \cc) = \E_z \left[ \sum_{i=1}^{D} \sum_{j=1}^K h_{ij}(\cc) \left( (x_{ij} - z_i)^2 + \frac{j-1}{K} \right) \right] \enspace,
\end{equation*}
where the underlying distribution of $z$ is $\mathcal{N}(0, K^{-2} I)$. This function switches the active variables in $\xx$ by the categorical variables $\cc$. The global optima locate at $h_i(\cc) = [1, 0, \dots, 0]$ for $i=1, \dots, D$, $\xx_1 = [0, \dots, 0]$, and arbitrary variables of $\xx_i$ for $i=2, \dots, D$. To mimic NN training, we approximate the expectation by using a sample $\bm{z}$, which is drawn from $\mathcal{N}(0, K^{-2} I)$ every parameter update.

We use the stochastic gradient descent (SGD) with a momentum of $0.9$ to optimize $\xx$ and anneal the step-size $\epsilon_{\xx}$ by cosine scheduling \citep{Loshchilov2017}, which we also use for the latter experiments. We initialize $\xx \sim \mathcal{N}(0, I)$ and the distribution parameter $\theta = (1/K)\mathbf{1}$. We regard it successfully optimized if a solution with the actual objective value less than $K^{-1} + D K^{-2}$ is sampled before $10^5$ iterations. We report the number of iterations to sample the target value divided by the success rate over 100 runs as the performance measure. We have tested different combinations of $D$ and $K$ and observed similar results. We report the results for $D = 30$ and $K=5$ as a typical one.
Figure \ref{fig:benchmark} compares ASNG, SNG (stochastic natural gradient with constant step-size), and Adam \citep{Kingma2015}---a standard step-size adaptation for DNNs---with different initial step-size. We replace the gradient in Adam with the normalized natural gradient as it is used in ASNG since we found in our preliminary studies that Adam does not work properly with its default. For SNG and Adam one needs to fine tune the step-size, otherwise they fail to locate the optimum. On the other hand, ASNG relaxes the sensitivity against $\delta^0_\theta$. The robustness of ASNG on the choice of $\alpha$ is evaluated in the supplementary material.

\begin{figure*}[tb]
\def\@captype{table}
\begin{minipage}[c]{.70\textwidth}
	\centering
	\tblcaption{Comparison of different architecture search methods on CIFAR-10. The search cost indicates GPU days for architecture search excluding the retraining cost.}
	\label{tbl:img-class}
    \begin{tabular}{lccccc}
    	\toprule
    	Method & Search Cost & Params & Test Error\\
    	 & (GPU days) & (M) & (\%)\\
    	\midrule
    	NASNet-A \citep{Zoph2018} & $1800$      & $3.3$    & $2.65$\\
    	NAONet \citep{Luo2018} & $200$ & $128$ & $2.11$\\
      \midrule
    	ProxylessNAS-G \citep{Cai2019} & $4$ & $5.7$ & $2.08$\\ 
    	SMASHv2 \citep{Brock2018}                     & $1.5$        & $16.0$   & $4.03$\\    
      \midrule  
    	DARTS second order \citep{Liu2019} & $4$ & $3.3$ & $2.76~(\pm 0.09)$ \\
    	DARTS first order \citep{Liu2019} &$1.5$ & $3.3$ & $3.00~(\pm 0.14)$ \\
    	SNAS  \citep{Xie2019} & $1.5$ & $2.8$ & $2.85~(\pm 0.02)$\\
    	ENAS \citep{Pham2018}            & $0.45$       & $4.6$    & $2.89$\\
    	\proposed & $0.11$ & $3.9$ & $2.83~(\pm 0.14)$\\
    	\bottomrule
    \end{tabular}
\end{minipage} 
\hfill %
\begin{minipage}[c]{.29\textwidth}
	\centering
	\includegraphics[width=0.78\hsize]{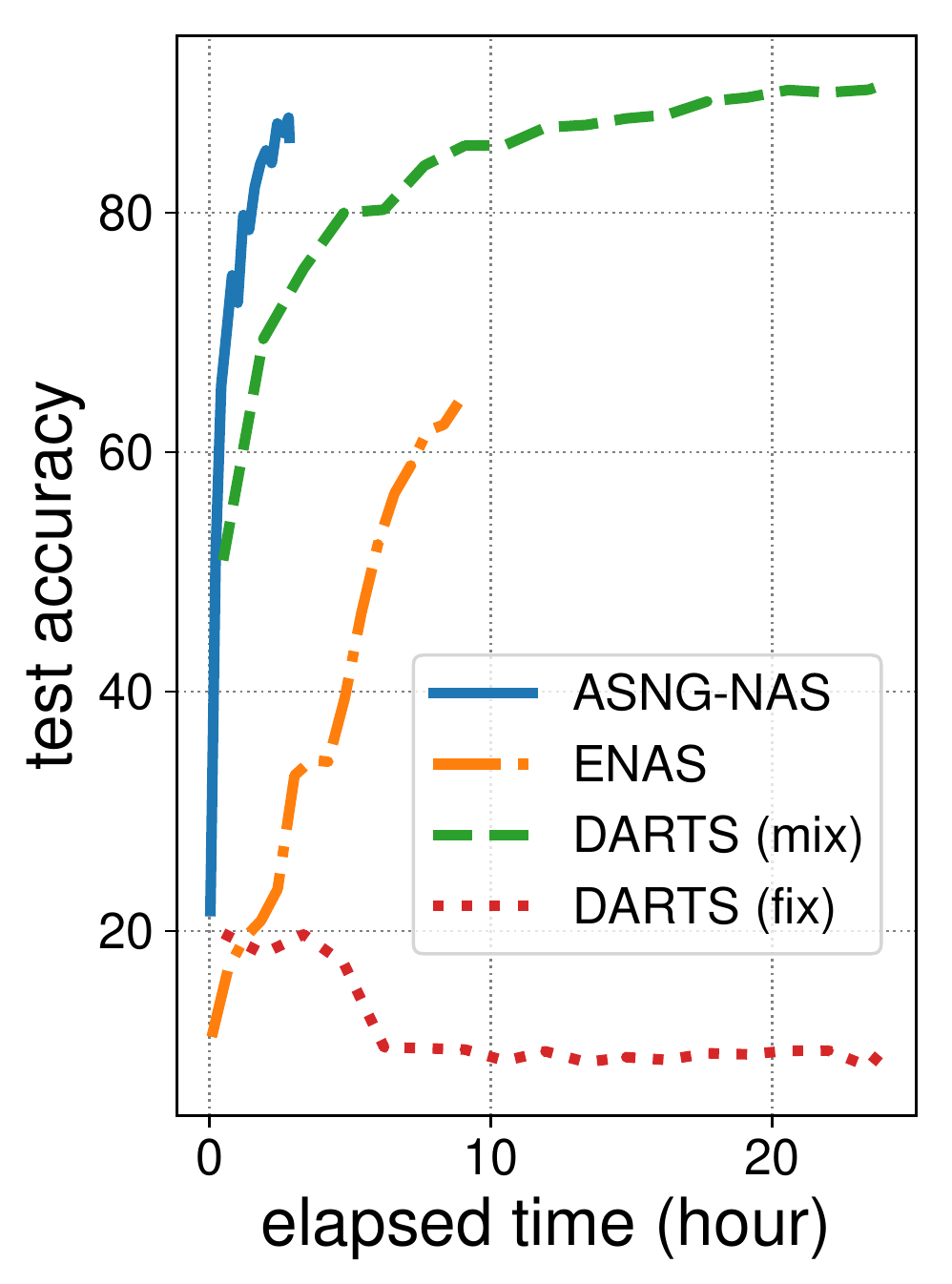}%
	\caption{Transitions of test error against elapsed time in the architecture search phase.
	}
	\label{fig:test_ac_hour}
\end{minipage}
\end{figure*}

\subsection{Architecture Search for Image Classification}
\label{sec:exp_class}

\paragraph{Dataset:}
We use the CIFAR-10 dataset and adopt the standard preprocessing and data augmentation as done in the previous works, e.g., \citet{Liu2019,Pham2018}.
During the architecture search, we split the training dataset into halves as $\Dataset = \{ \Dataset_{\xx}, \Dataset_{\theta} \}$ as done in \citet{Liu2019}. The gradients \eqref{eq:gx} and \eqref{eq:gt} are calculated using mini-batches from $\Dataset_{\xx}$ and $\Dataset_{\theta}$, respectively. We use the same mini-batch samples among the different architecture parameters in \eqref{eq:gx} and \eqref{eq:gt} to get accurate gradients. Note that we do not need the back-propagation for calculating \eqref{eq:gt}. Namely, the computational cost of the $\theta$ update is less than that of $\xx$.

\paragraph{Search Space:}
The search space is based on the one in \citet{Pham2018}, which consists of models obtained by connecting two motifs (called normal cell and reduction cell) repeatedly. Each cell consists of $B ~(= 5)$ nodes and receives the outputs of the previous two cells as inputs. Each node receives two inputs from previous nodes, applies an operation to each of the inputs, and adds them. Our search space includes 5 operations: identity, $3 \times 3$ and $5 \times 5$ separable convolutions \citep{Chollet2017}, and $3 \times 3$ average and max poolings. The separable convolutions are applied twice in the order of ReLU-Conv-BatchNorm. We select a node by 4 categorical variables representing 2 outputs of the previous nodes and 2 operations applied to them. Consequently, we treat $4B$-dimensional categorical variables for each cell. After deciding $B$ nodes, all of the unused outputs of the nodes are concatenated as the output of the cell. The number of the categorical variables is $n_{\cc} = 40$, and the dimension of $\theta$ becomes $n_\theta = 140$.

\paragraph{Training Details:}
In the architecture search phase, we optimize $\xx$ and $\theta$ for 100 epochs (about 40K iterations) with a mini-batch size of $64$. We stack 2 normal cells ($N = 2$) and set the number of channels at the first cell to 16. We use SGD with a momentum of 0.9 to optimize weights $\xx$. The step-size $\epsilon_{\xx}$ changes from 0.025 to 0 following the cosine schedule \citep{Loshchilov2017}. After the architecture search phase, we \emph{retrain} the network with the most likely architecture, $\hat{\cc} = \argmax_{\cc} p_\theta (\cc)$, from scratch, which is a commonly used technique \citep{Brock2018,Liu2019,Pham2018} to improve final performance. In the retraining stage, we can exclude the redundant (unused) weights. Then, we optimize $\xx$ for 600 epochs with a mini-batch size of $80$. We stack 6 normal cells ($N = 6$) and increase the number of channels at the first cell so that the model has the nearly equal number of weight parameters to 4 million. We report the average (avg.) and standard deviation (std.) among 3 independent experiments.

\paragraph{Result and Discussion:}
Table~\ref{tbl:img-class} compares the search cost and the test error of different NAS methods. The bottom 5 methods adopt similar search spaces, hence showing the performance differences due to search algorithms. The avg.\ and std.\ of \proposed are those of architecture search + retraining (whole NAS process), whereas the values for the other methods are taken from the references and have different meanings. E.g., the values for DARTS and SNAS are the avg. and std. of 10 independent \emph{retraining} of the best found architecture among 4 NAS processes. 

We clearly see the trade-off between search cost and final performance. The more accurate the performance estimation of neural architecture is (as in NASNet and NAONet), the better final performance is obtained at the risk of speed. Among relatively fast NAS methods (ENAS, DARTS, SNAS, and \proposed[false]), \proposed is the fastest and achieves a competitive error rate. The reason of speed difference between these algorithms is discussed in \S\ref{sec:dis}. 
We observed that the probability vector $\theta$ of the categorical distribution converges to a certain category. More precisely, the average value of the $\max_j [\theta]_{i, j}$ reaches around $0.9$ at the $50$th epoch. The architecture of the best model obtained by \proposed is found in supplementary material.

Figure~\ref{fig:test_ac_hour} compares the test error \wrt elapsed time in the architecture search phase. The test accuracy of the most likely architecture $\hat{\cc}$ is plotted for \proposed[false]. DARTS (mix) and DARTS (fix) are the architectures obtained by the same run of DARTS. The former is the one mixing all possible operations with real-valued structure parameters and is the one optimized during the architecture search, whereas the latter is the one that takes the operations with the highest weights and is the one used after the architecture search. We see that DARTS (fix) do not improve the test accuracy during the architecture search phase and the retraining is a must. DARTS (mix) achieves better performance than \proposed[false] in the end, but the obtained architecture is not one-hot and is computationally expensive. \proposed[false] shows the best performance for small time budgets.

\subsection{Architecture Search for Inpainting}
\label{sec:exp_inpaint}

\paragraph{Dataset:}
We use the CelebFaces Attributes Dataset (CelebA) \citep{Liu2015}. The data preprocessing and augmentation method is the same as \citet{Suganuma2018}.
We use three different masks to generate images with missing regions; a central square
block mask (Center); a random pixel mask where 80\% of the pixels were randomly masked (Pixel), and a half image mask where either the vertical or horizontal half of the image is randomly selected (Half).
Following \citet{Suganuma2018}, we use two standard evaluation measures: the peak-signal to noise ratio (PSNR) and the structural similarity index (SSIM) \citep{Wang2004} to evaluate the restored images. Higher values of these measures indicate a better image restoration.

\paragraph{Search Space:}
The search space we use is based on \citet{Suganuma2018} for comparison. The architecture encoding is slightly different but it can represent the exact same network architectures. We employ the symmetric convolutional autoencoder (CAE) as a base architecture. A skip connection between the convolutional layer and the mirrored deconvolution layer can exist. We prepare six types of layers: the combination of the kernel sizes $\{ 1 \times 1, 3 \times 3, 5 \times 5 \}$ and the existence of the skip connection. The layers with different settings do not share weight parameters.

We implement two \proposed[false] algorithms with only categorical variables (\proposed[false] (Cat)) and with mixed categorical and ordinal (integer) variables (\proposed[false] (Int)) to demonstrate the flexibility of the proposed approach. The former encodes the layer type, channel size, and connections for each hidden layer, and the connection for the output layer using categorical variables. We select the output channel size of each of $20$ hidden layers from $\{ 64, 128, 256 \}$. The latter encodes the kernel size and the channel size by integers in $\llbracket 1, 3 \rrbracket$ (corresponding to $\{ 1 \times 1, 3 \times 3, 5 \times 5 \}$) and $\llbracket 64, 256 \rrbracket$. We employ the Gaussian distribution as described in \S\ref{sec:asngnas}. Sampled variables are clipped to $[1, 3]$ and $[64, 256]$ and rounded to integers (only for architecture evaluation). The dimension of $\theta$ amounts to $n_\theta = 214$ for \proposed[false] (Cat) and $n_\theta = 174$ for \proposed[false] (Int). 

\begin{table*}[tb]
\caption{Results on the inpainting tasks. CE and SII indicate the context encoder \citep{Pathak2016} and the semantic image inpainting \citep{Yeh2017}, which are the human-designed CNN. E-CAE refers to the model obtained by the architecture search method using the evolutionary algorithm \citep{Suganuma2018}. BASE is the same depth of the best architecture obtained by E-CAE but having 64 channels and $3 \times 3$ filters in each layer, along with a skip connection. \proposed (Cat) encodes all architecture parameters into categorical variables, whereas \proposed (Int) encodes the kernel and channel sizes into integer variables. The values of CE, SII, BASE, and E-CAE are referenced from \citet{Suganuma2018}.}

\label{tbl:img-inpaint}
\begin{center}
\begin{tabular}{ccccccc}
	\toprule
               & \multicolumn{5}{c}{PSNR [dB] / SSIM}       \\
		           \cmidrule(r){2-7}
    \multirow{2}{*}{Mask} &  \multirow{2}{*}{CE} & \multirow{2}{*}{SII} & \multirow{2}{*}{BASE} & E-CAE & \proposed (Cat) & \proposed (Int) \\
     & & & & (12 GPU days) & (0.84 GPU days) & (0.75 GPU days) \\
	\midrule
    Center & $28.5$ / $0.912$ & $19.4$ / $0.907$ & $27.1$ / $0.883$ & \bm{$29.9$} / \bm{$0.934$} & $29.2$ / $0.903$ & $29.3$ / $0.911$ \\     
	Pixel    & $22.9$ / $0.730$ & $22.8$ / $0.710$ & $27.5$ / $0.836$ & $27.8$ / $0.887$ & $28.4$ / $0.905$ & \bm{$28.6$} / \bm{$0.909$} \\    
	Half     & $19.9$ / $0.747$ & $13.7$ / $0.582$ & $11.8$ / $0.604$ & \bm{$21.1$} / $0.771$ & $20.5$ / \bm{$0.779$} & $20.6$ / \bm{$0.779$} \\    
    \bottomrule
\end{tabular}
\end{center}
\end{table*}

\paragraph{Training Details:}
We use the mean squared error (MSE) as the loss function and a mini-batch size of $16$. In the architecture search phase, we use SGD with momentum with the same setting in \S\ref{sec:exp_class}, while we use Adam in the retraining phase. We apply gradient clipping with the norm of 5 to prevent a too long gradient step. The maximum numbers of iterations are $50K$ and $500K$ in the architecture search and retraining phases, respectively. The setting of the retraining is the same as in \citet{Suganuma2018}. Differently from the previous experiment, we retrain the obtained architecture without any change in this experiment.

\paragraph{Result and Discussion:}
Table~\ref{tbl:img-inpaint} shows the comparison of PSNR and SSIM.
The performances of \proposed are better than CE, SII, and BASE on all mask types and comparable to E-CAE.
\citet{Suganuma2018} reported that E-CAE spent approximately 12 GPU days (3 days with 4 GPUs) for the architecture search and retraining. 
On the other hand, the average computational times of \proposed were less than 1 GPU days. \proposed (Cat) took approximately 6 hours for the architecture search and 14 hours for the retraining on average, whereas the average retraining time of \proposed (Int) was reduced to 11 hours. This is because the architectures obtained by \proposed (Int) tended to have a small number of channels compared to \proposed (Cat) that selects from the predefined three channel sizes.

In conclusion, \proposed achieved practically significant speedup over E-CAE without compromising the final performance. The flexibility of \proposed has been shown as well. The capability of \proposed to treat mixed categorical and ordinal variables potentially decreases the number of the architecture parameters (good for speed) and enlarges the search space (good for performance).

\section{Related Work and Discussion}
\label{sec:dis}

\proposed falls into one-shot NAS. On this line of the research, three different existing approaches are reported. The 1st category is based on a meta-network. SMASH \citep{Brock2018} employs HyperNet that takes an architecture $\cc$ as its input and returns the weights for the network with architecture $\cc$. The weights of HyperNet is then optimized by backprop while $\cc$ is randomly chosen during architecture search. ENAS \citep{Pham2018} employs a recurrent neural network (RNN) to generate a sequence of categorical variables $\cc$ representing neural architecture. It optimizes the weights and the RNN weights alternatively. ENAS and \proposed are different in that the latter directly introduces a probability distribution behind $\cc$ while the former employ RNN. The advantage of \proposed over meta-network based approaches is that we do not need to design the architecture of a meta-network, which may be a tedious task for practitioners.  

The 2nd category is based on continuous relaxation. DARTS \citep{Liu2019} extends essentially categorical architecture parameters (selection of operations and connections) to a real-valued vector by considering a linear combination of outputs of all possible operations. This enables gradient descent both on the connection weights and the weights for the linear combination. This seminal work is followed by further improvements \cite{Xie2019,Cai2019}. An advantage of \proposed is that DARTS requires to compute all possible operations and connections to perform backprop, whereas we only require to process sub-networks with sampled architectures $\cc$, hence \proposed is faster. This advantage is reflected in Table~\ref{tbl:img-class}. Another advantage is its flexibility in the sense that the continuous relaxation of DARTS requires the output of all possible operations to live in the same domain to add them. 

The last category is based on stochastic relaxation, which is another approach enabling to use gradient descent. \citet{Shirakawa2018} has introduced it to model connections and types of activation functions in multi-layer perceptrons. They are encoded by a binary vector $\cc$ and Bernoulli distribution is considered as the underlying distribution of $\cc$. The probability parameters of Bernoulli distribution is updated by SNG. We improve their work in the following directions: generalization to arbitrary architecture parameters (categorical, ordinal, or their mixture), theoretical investigation of monotone improvement, robustness against its input parameter by introducing a step-size adaptation mechanism.

This paper focused on the optimization framework for NAS. One can easily incorporate a different search space and a different performance estimation method into our framework. The step-size adaptation mechanism eases hyper-parameter tuning when different components are introduced. The ability to treat ordinal variables such as the number and size of filters and the number of layers accepts more flexible search space. In existing studies they are modeled by categorical variables by choosing a few representative numbers beforehand. Moreover, the ordinal variables potentially decreases the dimension of architecture parameters. When multiple GPUs are available, \proposed can easily enjoy them by increasing $\lambda_{\xx}$ and $\lambda_{\cc}$ and distributing them. In our preliminary study, we found that the larger they are, the greater step-size are allowed and the step-size adaptation automatically increases it. Our simple formulation allows theoretical investigation, which we think is missing in the current NAS research fields. Further theoretical investigation will contribute better understanding and further improvement of NAS. 

One-shot NAS including our method does not optimize parameters involved in learning process such as the step-size for weight update. It is because their effects do not appear in the one-shot loss and will not be optimized effectively. If we employ hyper-parameter optimizers such as Bayesian optimization to optimize these parameters while each training process is replaced by our method, both architectures and other hyper-parameters could be optimized. The fast and robust properties of our method will be useful to combine one-shot NAS and hyper-parameter optimizer. This is an important direction towards automation of deep learning.

\section*{Acknowledgement}
This work is partially supported by the SECOM Science and Technology Foundation.

\bibliography{bibliography}
\bibliographystyle{icml2019}

\clearpage
\input{supplemental_body}

\end{document}

%% file: supplemental_body.tex
\appendix

\section{Proof of Proposition~3}\label{apdx:proof}
\begin{proof}It follows
  \begin{align*}
    \MoveEqLeft[1]\ln J(\theta') - \ln J(\theta) \\
 &= \ln \int \frac{ p_{\theta'}(\cc) }{ p_{\theta}(\cc) } \frac{ f(\cc) }{ J(\theta) } p_{\theta}(\cc) \\
 &\geq \int \ln \left( \frac{ p_{\theta'}(\cc) }{ p_{\theta}(\cc) }\right) \frac{ f(\cc) }{ J(\theta) } p_{\theta}(\cc) \\
 &\geq \int_{\cc:  p_{\theta'}(\cc) < p_{\theta}(\cc) } \ln \left( \frac{ p_{\theta'}(\cc) }{ p_{\theta}(\cc) }\right) \frac{ f(\cc) }{ J(\theta) } p_{\theta}(\cc)\\
 &\geq \frac{ f^* }{ J(\theta) }\int_{\cc:  p_{\theta'}(\cc) < p_{\theta}(\cc) } \ln \left( \frac{ p_{\theta'}(\cc) }{ p_{\theta}(\cc) }\right)  p_{\theta}(\cc)\\
 &= - \frac{ f^* }{ J(\theta) }\int_{\cc:  p_{\theta'}(\cc) < p_{\theta}(\cc) } \ln \left( \frac{ p_{\theta}(\cc) }{ p_{\theta'}(\cc) }\right)  p_{\theta}(\cc)\\
 &\geq - \frac{ f^* }{ J(\theta) } D_\theta(\theta', \theta  ) \enspace.\qedhere
  \end{align*}
\end{proof}

\section{Derivation for Categorical Distribution}

We derive the Fisher information matrix, its inverse and square root, and the natural gradient for the categorical distribution defined on $\mathcal{C} = \llbracket 1,m_1\rrbracket\times \cdots \times\llbracket1, m_{n_{\cc}}\rrbracket$.

In our parameterization $\theta = (\theta_1, \dots, \theta_{n_{\cc}})$, the probability of $i$-th ($i \in \llbracket 1, n_{\cc}\rrbracket$) categorical variable $[\cc]_{i}$ to be $j \in \llbracket 1, m_i-1\rrbracket$ is $[\theta]_{i,j} = [\theta_i]_{j}$, and $1 - \sum_{j=1}^{m_i-1} [\theta]_{i,j}$ is the probability of $[\cc]_{i} = m_i$. For the sake of simplicity, we denote $[\theta]_{i,m_i} = 1 - \sum_{j=1}^{m_i-1} [\theta]_{i,j}$. Then, $T(\cc) = (T_1([\cc]_1), \dots, T_{n_{\cc}}([\cc]_{n_{\cc}}))$, where $T_i: \llbracket 1, m_{i}\rrbracket \to [0, 1]^{m_i-1}$ is the one-hot representation without the last element. It is the exponential parameterization as $\theta = \E[T(\cc)]$. 

The inverse of the Fisher information matrix simply follows from the formula for the exponential family: $\Fi(\theta)^{-1} = \E[(T(\cc) - \theta)(T(\cc) - \theta)^\T]$. It is a block diagonal matrix $\Fi(\theta)^{-1} = \diag(\Fi_1(\theta_1)^{-1}, \dots, \Fi_{n_{\cc}}(\theta_{n_{\cc}})^{-1})$, where $\Fi_i(\theta_i)^{-1} = \diag(\theta_i) - \theta_i \theta_i^\T$. Sherman-Morrison formula reads
$\Fi_i(\theta_i) = \diag(\theta_i)^{-1} + (1 - \sum_{j=1}^{m_i-1} [\theta_i]_{j})^{-1}\bm{1}\bm{1}^\T$ and we have $\Fi(\theta) = \diag(\Fi_1(\theta_1), \dots, \Fi_{n_{\cc}}(\theta_{n_{\cc}}))$. 

As the Fisher information matrix is block-diagonal, and each block is of size $m_i - 1$, a naive computation of $\Fi(\theta)^\frac12$ requires $O(\sum_{i=1}^{n_{\cc}} (m_i - 1)^3)$. This is usually not expensive as $n_{\cc} \gg m_i$. An alternative way that we employ in this paper is to replace $\Fi(\theta)^\frac12$ with a tractable factorization $A$ with $\Fi(\theta) = A A^\T$. Our choice of $A$ is the block-diagonal matrix whose $i$-th block is square, of size $m_i - 1$, and 
\begin{equation*}
  A_i = \mathrm{diag}(\theta_i)^{-\frac12}  + \frac{1}{ \sqrt{[\theta]_{i, m_i}} + [\theta]_{i, m_i} } \bm{1} \sqrt{\theta_i}^\T \enspace,
\end{equation*}
where $\sqrt{\theta_i}$ is a vector whose $j$-th element is the square root of $[\theta]_{i, j}$. Then, the product of $A$ and a vector can be computed in $O(\sum_{i=1}^{n_{\cc}} (m_i - 1))$. In our preliminary study we did not obverse any significant performance difference by this approximation.

\section{Derivation for Gaussian Distribution}

We derive the Fisher information matrix, its inverse and square root, and the natural gradient for the Gaussian distribution defined on $\mathcal{C} \subseteq \mathbb{R}^{n_{\cc}}$. 

Our choice is $P_\theta = \mathcal{N}(\mu_1, \sigma_1^2) \times \cdots \times\mathcal{N}(\mu_{n_{\cc}}, \sigma_{n_{\cc}}^2)$ and $\theta = (\mu_1, \mu_1^2 + \sigma_1^2, \dots, \mu_{n_{\cc}}, \mu_{n_{\cc}}^2 + \sigma_{n_{\cc}}^2)$. Then, we have $T(\cc) = (T_1([\cc]_{1}),\dots, T_{n_{\cc}}([\cc]_{n_{\cc}}))$ with $T_i([\cc]_i) = ([\cc]_{i}, [\cc]_{i}^2)$. It is the exponential parameterization as $\theta_{i} = (\mu_i, \mu_i^2+\sigma_i^2) = \E[T_i([\cc]_i)]$ and $\theta = (\theta_1, \dots, \theta_{n_{\cc}})$. 

The inverse of the Fisher information matrix simply follows from the formula for the exponential family. $\Fi(\theta)^{-1}$ is a block-diagonal matrix with block size $2$ whose $i$-th block is $\Fi_i(\theta_i)^{-1} = [\sigma_i^2, 2\mu_i\sigma_i^2; 2\mu_i\sigma_i^2, 4\mu_i^2\sigma_i^2 + 2\sigma_i^4]$. Since each block is a symmetric matrix of dimension $2$, its eigen decomposition $\Fi_i(\theta_i)^{-1} = \mathbf{E}\Lambda\mathbf{E}^\T$ can be analytically obtained. With the decomposition, we have $\Fi_i(\theta_i) = \mathbf{E}\Lambda^{-1}\mathbf{E}^\T$ and $\Fi(\theta_i)^{\frac12} = \mathbf{E}\Lambda^{-\frac12}\mathbf{E}^\T$. Then, we have $\Fi(\theta) = \diag(\Fi_1(\theta_1), \dots, \Fi_{n_{\cc}}(\theta_{n_{\cc}}))$ and $\Fi(\theta)^{\frac12} = \diag(\Fi_1(\theta_1)^{\frac12}, \dots, \Fi_{n_{\cc}}(\theta_{n_{\cc}})^{\frac12})$.

\section{Restriction for the Range of $\theta$}
\label{sec:theta_restriction}

To guarantee that the Fisher information matrix is nonsingular and the natural gradient is well defined, we restrict the domain $\Theta$ of the parameter of the probability distribution. 

For the categorical distribution, we set $\Theta = [ \theta_{1}^{\min}, \theta^{\max} \big]^{m_1-1} \times \cdots \times [ \theta_{n_{\cc}}^{\min}, \theta^{\max} ]^{m_{n_{\cc}}-1}$, where $\theta_{i}^{\min} = \frac{1}{n_{\cc} (m_i - 1)}$ and $\theta^{\max} = 1 - \frac{1}{n_{\cc}}$. Then, a small yet positive probability for all combinations of categorical variables is guaranteed and the Fisher information matrix is nonsingular at any point of $\Theta$. 

To force the parameter to live in $\Theta$, we apply the following steps after $\theta$ update:
\begin{align*}
[\theta]_{i, j} &\leftarrow \max\{ [\theta]_{i, j}, \theta_{i}^{\min} \} \text{ for all $i, j$, then} \\
[\theta]_{i,j} &\leftarrow [\theta]_{i,j} + \frac{1 - \sum^{m_i}_{k=1} [\theta]_{i, k}}{\sum^{m_i}_{k=1} \left( [\theta]_{i, k} - \theta_{i}^{\min} \right)} \left( [\theta]_{i,j} - \theta_{i}^{\min} \right) \enspace.
\end{align*}
The first line guarantees $[\theta]_{i,j} \geq \theta_{i}^{\min}$.  The second line ensures $\sum_{j=1}^{m_i} [\theta]_{i,j} = 1$, while keeping $[\theta]_{i,j} \geq \theta_{i}^{\min}$.

For the integer variables, the parameters of the Gaussian distributions, $[\theta]_{i, 1} := \mu_i$ and $[\theta]_{i, 2} := \mu_i^2 + \sigma_i^2$, are forced to be in a compact set as follows. The range of the mean value of each integer variable is $[ \mu_i^{\min}, \mu_i^{\max}]$, which is the same as the range of the integer variable. The standard deviation is forced to be no smaller than $\sigma_i^{\min} = 1/4$ and no greater than $\sigma_i^{\max} = (\mu_i^{\max} - \mu_i^{\min}) / 2$. To keep the parameters inside these ranges, after every $\theta$ update we clip $[\theta]_{i, 1}$ to $[ \mu_i^{\min}, \mu_i^{\max}]$ and $[\theta]_{i, 2}$ to $[[\theta]_{i, 1}^2 + (\sigma_i^{\min})^2, [\theta]_{i, 1}^2 + (\sigma_i^{\max})^2]$. If the variables are real-value, rather than integer, then $\sigma_i^{\min}$ may be set smaller depending on the meaning of the variable.

\section{Experimental Details}

\subsection{Toy Problem}
\label{sec:sup_test}
\begin{figure}[t]
  \centering%
  \begin{subfigure}{0.49\hsize}%
    \centering%
    \includegraphics[width=\hsize]{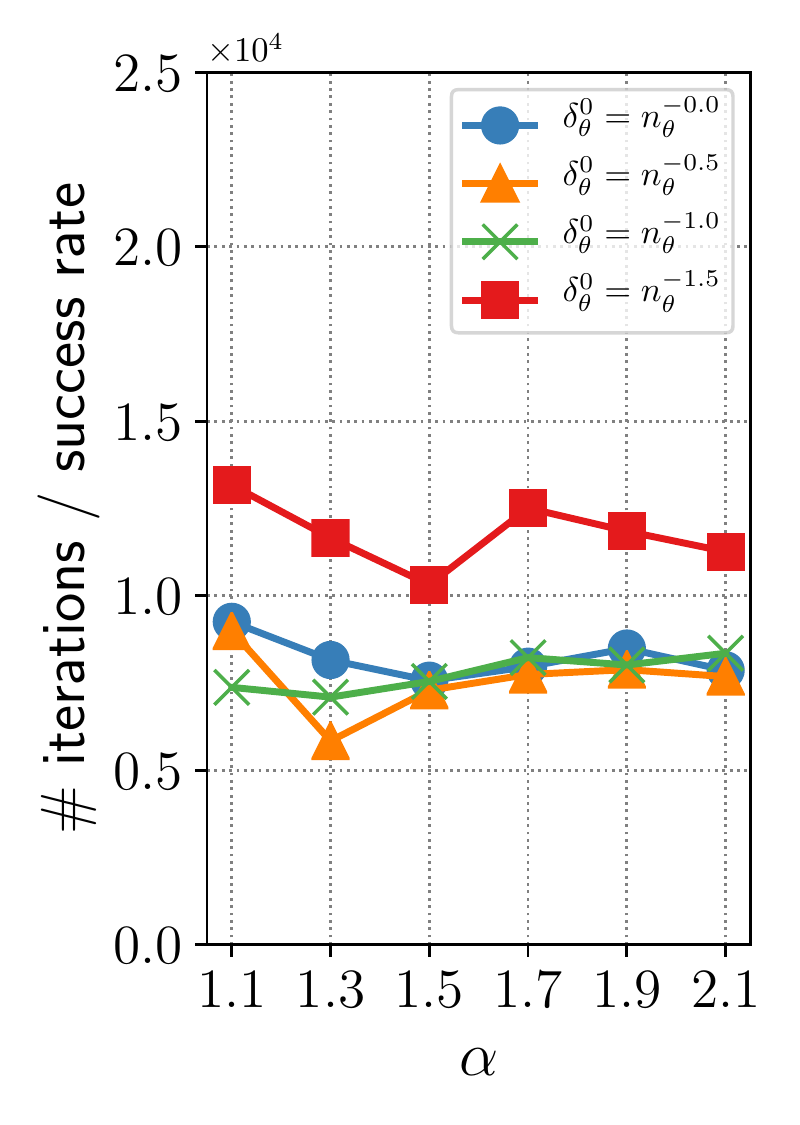}%
  \end{subfigure}%
  \begin{subfigure}{0.49\hsize}%
    \centering%
    \includegraphics[width=\hsize]{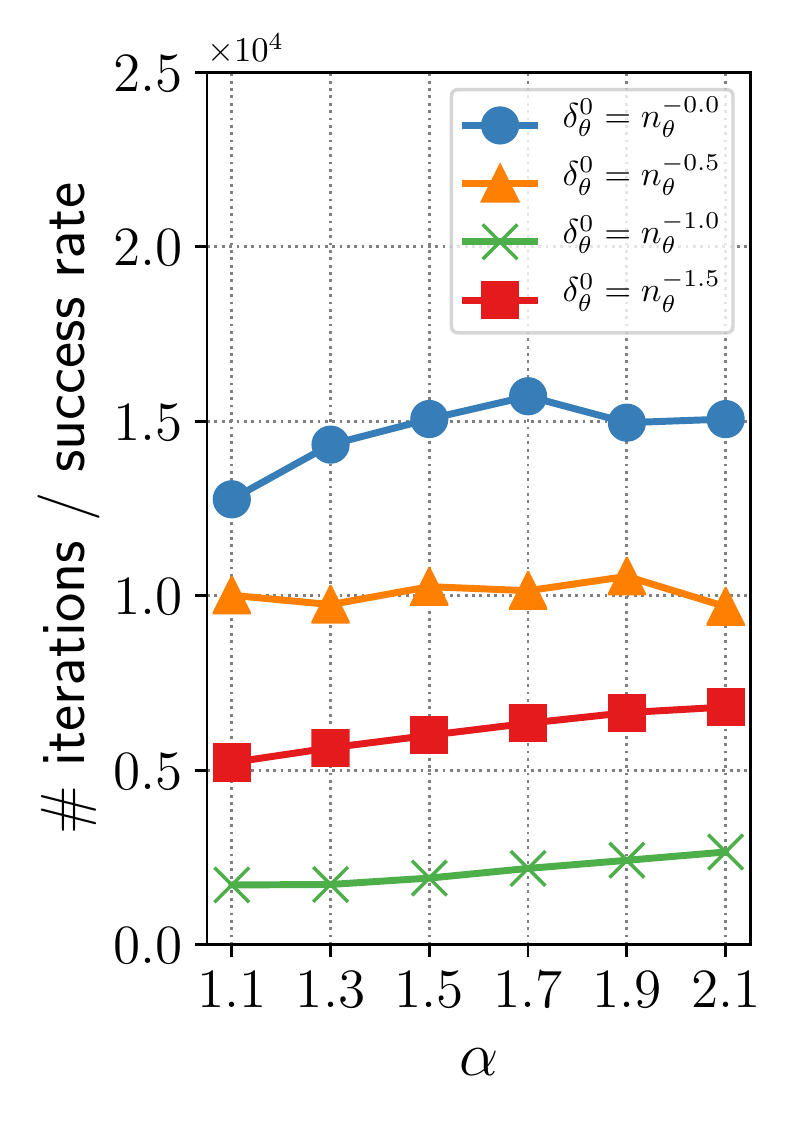}%
  \end{subfigure}%
  \caption{Performance of ASNG with the different $\alpha$ settings on the selective squared error function for $\epsilon_{\xx}=0.05$ (left) and $0.0005$ (right). Median values over $100$ runs are reported.}%
  \label{fig:benchmark_alpha}%
\end{figure}

To check the robustness of ASNG for the hyper-parameter $\alpha$, we ran ASNG on the selective squared error function with the varying $\alpha$ and initial step-size $\delta^0_\theta$ for the step-sizes of $\epsilon_{\xx} = \{0.05, 0.0005 \}$. Figure \ref{fig:benchmark_alpha} shows the performance of ASNG with the different $\alpha$ settings. We observe that the hyper-parameter $\alpha$ is not sensitive for the performance, and ASNG reaches the target value for all settings.

\subsection{Image Classification}
\label{sec:sup_class}

\paragraph{Dataset:}
We use the CIFAR-10 dataset which consists of 50,000 and 10,000 RGB images of 32 $\times$ 32, for training and testing. All images are standardized in each channel by subtracting the mean and then dividing by the standard deviation. We adopt the standard data augmentation for each training mini-batch: padding 4 pixels on each side, followed by choosing randomly cropped 32 $\times$ 32 images and by performing random horizontal flips on the cropped images. We also apply the cutout \citep{DeVries2017} to the training data.

\begin{figure}[t]
  \centering
  \includegraphics[width=0.99\linewidth]{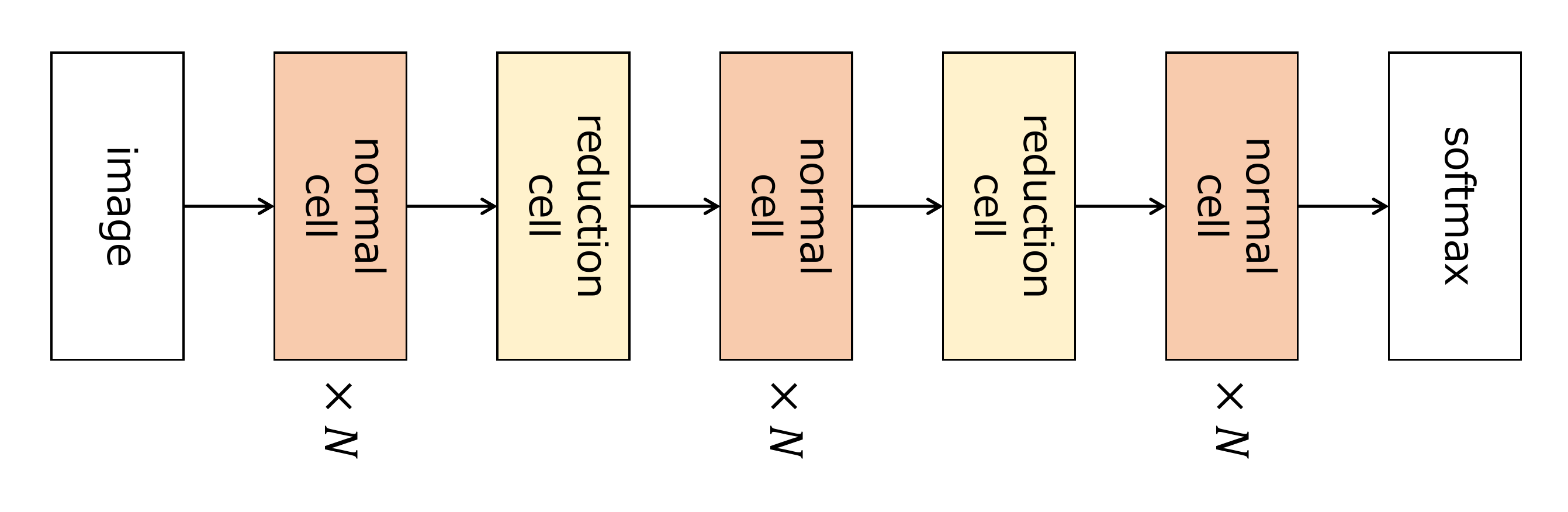}
  \caption{Overall model structure for the classification task.}
  \label{fig:nas-structure}
\end{figure}

\paragraph{Search Space:}
Figure \ref{fig:nas-structure} shows the overall model structure for the classification task. We optimize the architecture of the normal and reduction cells by \proposed[]. In the retraining phase, we construct the CNN using the optimized cell architecture with an increased number of cells $N$ and channels.
In the reduction cell, all operations applied to the inputs of the cell have a stride of 2, and the number of channels is doubled to keep the  dimension of the output roughly constant.

\begin{figure*}[!t]
  \centering
   \includegraphics[width=0.7\linewidth]{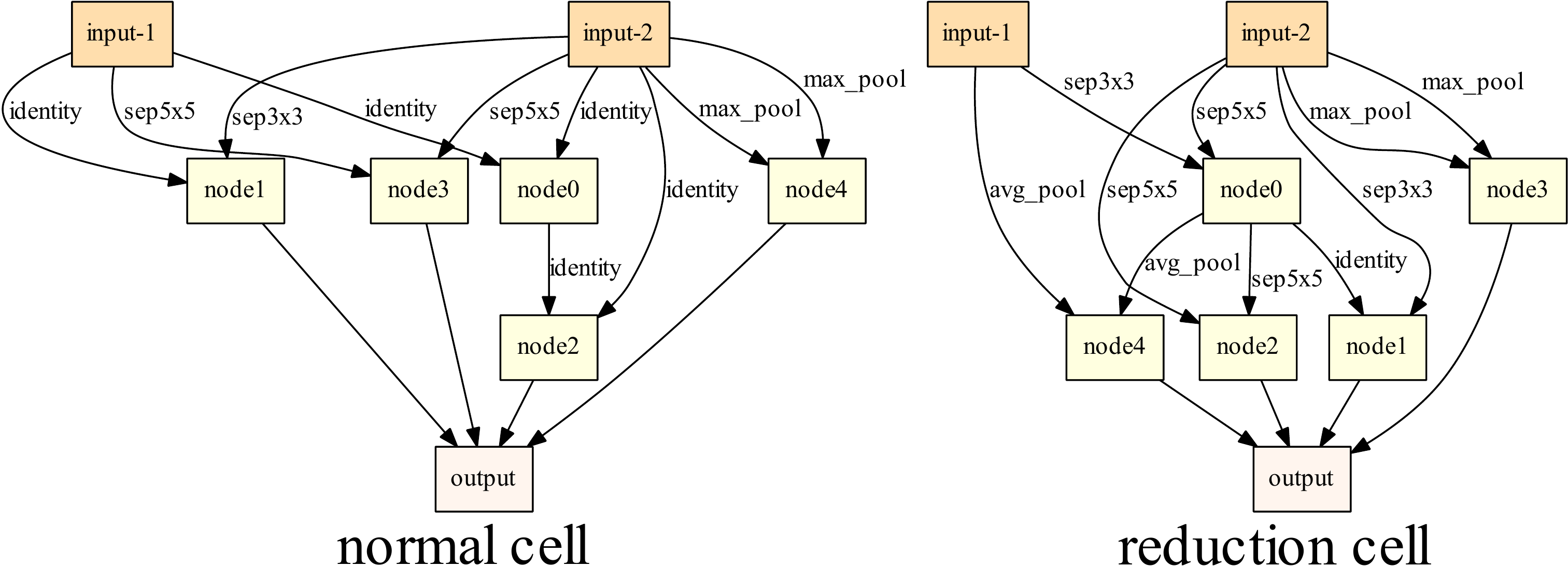}
  \caption{The best cell structures discovered by \proposed[false] in the classification task.}
  \label{fig:cell}
\end{figure*}

\paragraph{Training Details:}
In the architecture search phase, we fix affine parameters of batch normalizations for the purpose of absorbing effect of the dynamic change in architecture. We apply weight decay of $3 \times 10^{-4}$ and clip the norm of gradient at 5. 
In the retraining phase, we make all batch normalizations have learnable affine parameters because the architecture no longer changes. We apply the ScheduledDropPath \citep{Zoph2018} dropping out each path between nodes, and the drop path rate linearly increases from 0 to 0.3 during the training. We also add the auxiliary classifier \citep{Szegedy2016} with the weight of 0.4 that is connected from the second reduction cell. The total loss is a weighted sum of the losses of the auxiliary classifier and output layer. Other settings are the same as the architecture search phase.

\paragraph{The best cell structures:}
The best cell structures that achieve the error rate of $2.66 \%$ is displayed in Figure \ref{fig:cell}.

\begin{figure*}[!t]
  \centering
  \includegraphics[width=0.8\linewidth]{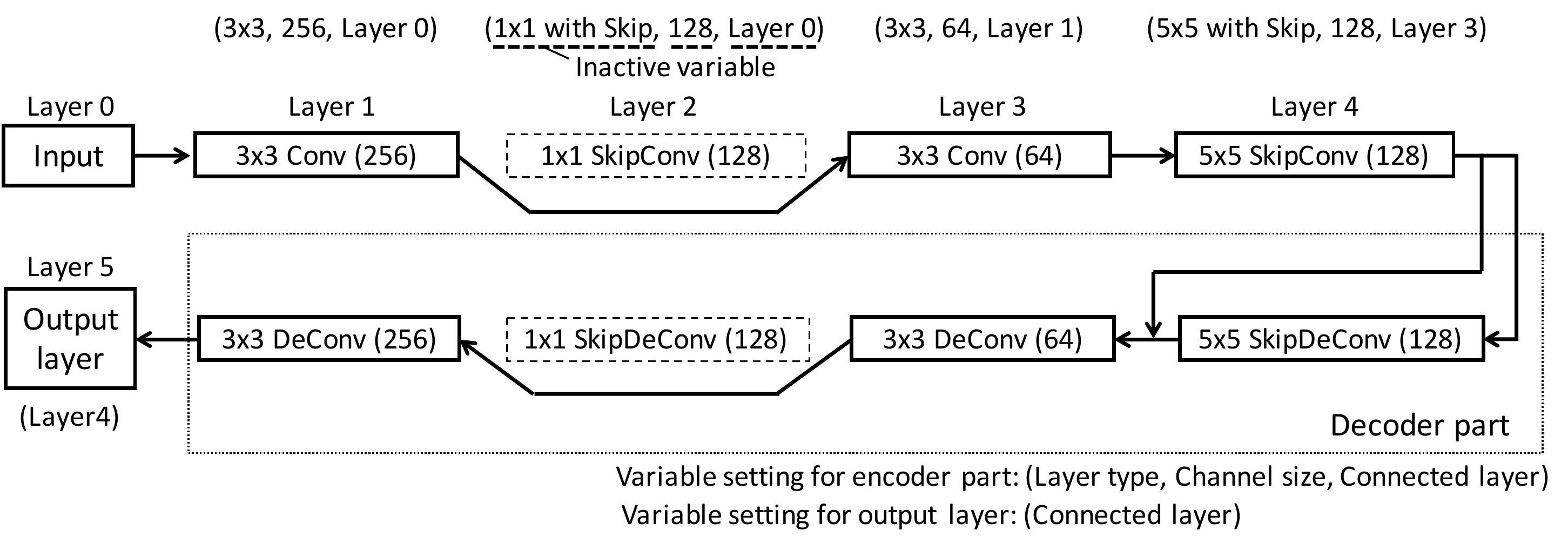}
  \caption{A conceptual example of the decoded symmetric CAE architecture and the corresponding categorical variables. The decoder part is automatically decided from the encoder structure as a symmetric manner.}
  \label{fig:cae}
\end{figure*}

\subsection{Inpainting}
\label{sec:sup_inpaint}
\paragraph{Dataset:}
The CelebA is a large-scale human face image dataset that contains 202,599 RGB images. We select 101,000 and 2,000 images for training and test, respectively, in the same way as \citet{Suganuma2018}. All images were cropped to properly contain the entire face by using the provided the bounding boxes, and resized to 64 $\times$ 64 pixels. All images are normalized by dividing by 255, and we perform data augmentation of random horizontal flipping on the training images.
We adopt three masks, Center, Pixel, and Half, to make corrupted images. The purpose of the task is to recover a clean image from the corrupted image as much as possible.
The masks in random pixel and half image masks were randomly generated for each training mini-batch and each test image.

\paragraph{Evaluation Measure:}
The PSNR is the metric evaluating the error between the ground truth and restored images and corresponds to the mean squared error (MSE). But the PSNR are not very well matched to perceived visual quality because the PSNR can not distinguish between the large difference on local region and the small difference on overall region. For this reason, the SSIM is also often used together with the PSNR, and more clearly assesses difference in each local region. We quantize the generated image within $[ 0, 255 ]$ followed by to calculate the PSNR and SSIM value. The setting of SSIM is based on \citet{Wang2004}.

\paragraph{Search Space:}
We employ the convolutional autoencoder (CAE), which is similar to RED-Net \citep{Mao2016}, as a base architecture. RED-Net consists of a chain of convolution layers and symmetric deconvolution layers as the encoder and decoder parts, respectively. The encoder and decoder parts perform the same counts of downsampling and upsampling with a stride of $2$, and a skip connection between the convolutional layer and the mirrored deconvolution layer can exist. For simplicity, each layer employs either a skip connection or a downsampling, and the decoder part is employed in the same manner. In the skip connected deconvolution layer, the input feature maps from the encoder part are added to the output of the deconvolution operation, followed by ReLU. In the other layers, the ReLU activation is performed after the convolution and deconvolution operations. We prepare six types of layers: the combination of the kernel sizes $\{ 1 \times 1, 3 \times 3, 5 \times 5 \}$ and the existence of the skip connection. The layers with different settings do not share weight parameters.

To represent a symmetric CAE, it is enough to represent the encoder part. We consider $N_\mathrm{c}$ hidden layers and the output layer. We encode the type, channel size, and connections of each hidden layer. The kernel size and stride of the output deconvolution layer are fixed with $3 \times 3$ and $1$, respectively, but the connection is determined by a categorical variable. To ensure the feed-forward architecture and to control the network depth, the connection of the $i$-th layer is only allowed to be connected from ($i-1$) to $\max (0, i-b)$-th layers, where $b$ ($b > 0$) is called the level-back parameter. Namely, the categorical variable representing the connection of the $i$-th layer has $\min (i, b)$ categories. Obviously, the first hidden layer always connects with the input, and we can ignore this part. With this representation, there can exist \textit{inactive} layers that do not connect to the output layer. Therefore, this model can represent variable length architectures by the fixed-dimensional variables. We choose $N_{c} = 20$ and the level-back parameter of $b=5$.

\proposed (Cat) encodes the type and channel size of each hidden layer by categorical variables with $6$ and $3$ categories, respectively. We select the output channel size of each hidden layer from $\{ 64, 128, 256 \}$. It amounts to $n_{\cc} = 60$ ($\#$ categorical variables) and $n_\theta = 214$ (dimension of $\theta$). A conceptual example of the symmetric CAE architecture and the corresponding representation by the categorical variables is shown in Figure \ref{fig:cae}. \proposed (Int) encodes the kernel size and the channel size by integers in $\llbracket 1, 3 \rrbracket$ (corresponding to $\{ 1 \times 1, 3 \times 3, 5 \times 5 \}$) and $\llbracket 64, 256 \rrbracket$. The existence of skip connection is determined by a categorical variable with $2$ categories. It amounts to $n_{\cc} = 80$ ($\#$ categorical and integer variables) and $n_\theta = 174$ (dimension of $\theta$).

\begin{figure}[tb]
  \centering
   \includegraphics[width=0.99\linewidth]{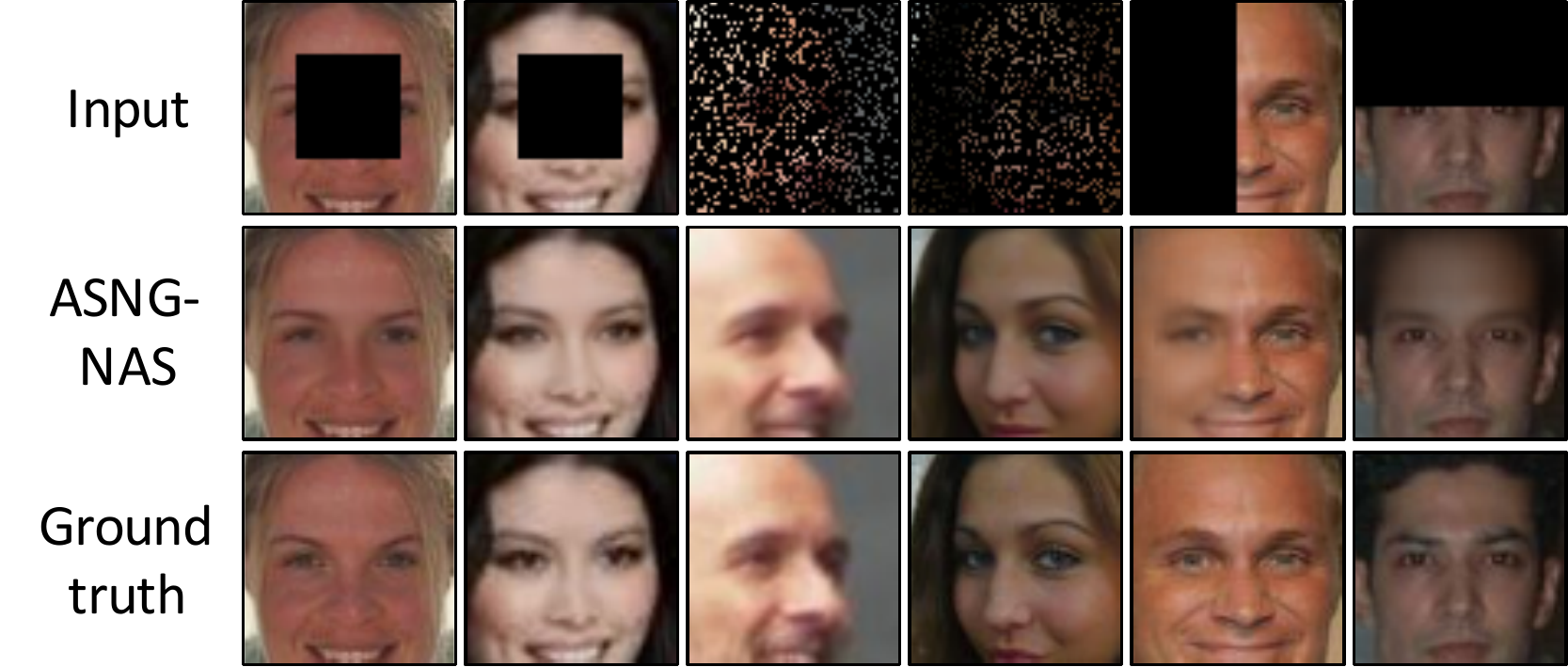}
  \caption{Example of inpainting results obtained by \proposed[false].}
  \label{fig:inpaint_out}
\end{figure}

\paragraph{Example of Inpainting Result:}
Figure \ref{fig:inpaint_out} shows the example of inpainting results obtained by \proposed[false].

